# Assessing a mobile-based deep learning model for plant disease surveillance

Amanda Ramcharan*[1], Peter McCloskey[1], Kelsee Baranowski[1], Neema Mbilinyi[2], Latifa Mrisho[2], Mathias Ndalahwa[2], James Legg[2], and David Hughes[3,4]

[1] Department of Entomology, College of Agricultural Sciences, Penn State University, State College, PA, United States
[2] International Institute of Tropical Agriculture, Dar es Salaam, Tanzania
[3] Department of Biology, Eberly College of Sciences, Penn State University, State College, PA, United States
[4] Center for Infectious Disease Dynamics, Huck Institutes of Life Sciences, Penn State University, State College, PA, United States

*Corresponding author. Email: amr418@psu.edu

# Abstract

Convolutional neural network models (CNNs) have made major advances in computer vision tasks in the last five years. Given the challenge in collecting real world datasets, most studies report performance metrics based on available research datasets. In scenarios where CNNs are to be deployed on images or videos from mobile devices, models are presented with new challenges due to lighting, angle, and camera specifications, which are not accounted for in research datasets. It is essential for assessment to also be conducted on real world datasets if such models are to be reliably integrated with products and services in society.
Plant disease datasets can be used to test CNNs in real time and gain insight into real world performance. We train a CNN object detection model to identify foliar symptoms of diseases (or lack thereof) in cassava (*Manihot esculenta* Crantz). We then deploy the model on a mobile app and test its performance on mobile images and video of 720 diseased leaflets in an agricultural field in Tanzania. Within each disease category we test two levels of severity of symptoms - mild and pronounced, to assess the model performance for early detection of symptoms. In both severities we see a decrease in the F-1 score for real world images and video.
The F-1 score dropped by 32% for pronounced symptoms in real world images (the closest data to the training data) due to a drop in model recall. If the potential of smartphone CNNs are to be realized our data suggest it is crucial to consider tuning precision and recall performance in order to achieve the desired performance in real world settings. In addition, the varied performance related to different input data (image or video) is an important consideration for the design of CNNs in real world applications.

# 1 Main

A landmark in computer vision occurred in 2012 when a deep convolutional neural network (CNN) won the Imagenet competition to classify over 1 million images from 1,000 categories, almost halving the error rates of its competition[1]. This success brought about a revolution in computer vision with CNNs dominating the approach for a variety of classification and detection tasks. Large tech companies and startups have capitalized on these advances to design real-time computer vision products and services while companies such as NVIDIA, Intel, Qualcomm and Samsung are developing CNN chips to enable real-time vision applications in smartphones, cameras, robots and self-driving cars [1].

As CNNs become the standard computer vision model to be deployed in real-time vision applications, assessing and reporting whether the results of their performance translates from research datasets to real time scenarios is crucial. Results of different CNN architectures are usually reported on standard large scale computer vision datasets of a million and more static images[2,3,4,5]. Domain specific datasets like medical imagery or plant diseases, where transfer learning is often applied to CNN models, comprise smaller datasets as expert labeled images are more challenging to acquire[6,7]. In a recent assessment for a skin lesion classification task, researchers reported the performance of the deep learning model matched at least 21 dermatologists tested across three critical diagnostic tasks[8]. This study was done on a labelled dataset of 129,450 clinical images and the researchers concluded that the technology is deployable on a mobile device but further evaluation in real-world settings is needed. Similar conclusions have been drawn by researchers[7,9,10,11,12]. Deploying on mobile devices would also be beneficial in democratizing access to algorithms while maintaining user privacy by running inference offline.

Despite the ubiquity of smartphones there are few examples of CNNs deployed on these phones categorizing visual scenes in the real world where performance is affected by input data type and compounded by wide extremes in lighting as is normal in outdoor settings. Clear examples of computer vision in real world settings such as autonomous vehicles (cars and drones) leverage multiple sensors in both the visible and non-visible spectrum[13,14]. If smartphone CNNs are to achieve their promise it is important to recognize the constraint of a single sensor (i.e. camera) and test the performance of CNNs on mobile devices in conditions they are intended to be used in.

Here, we investigate plant disease diagnostics on a mobile device. We deploy and test the performance of a CNN object detection model in a mobile app for real-time plant disease diagnosis in an agricultural field. Working with plant diseases provides a demanding case study because it is an outdoor setting with lighting conditions that could affect computer vision performance.

We use the Tensorflow platform to deploy a smartphone CNN object detection model designed to identify foliar symptoms of three diseases, two types of pest damage, and nutrient deficiency (or lack thereof) in cassava (*Manihot esculenta* Crantz). We utilize the Single Shot Multibox

(SSD) model with the MobileNet detector and classifier pre-trained on the COCO dataset (Common Objects in Context) of 1.5 million images (80 object categories). For simplicity, we refer to the CNN object detector model as the CNN model. We employ transfer learning to fine tune the model parameters to our dataset which comprised 2,415 cassava leaf images for pronounced symptoms of each class. The cassava leaf dataset was built with images taken in experimental fields of the International Institute of Tropical Agriculture (IITA), in Bagamoyo District, Tanzania. Complete details of this dataset were previously reported in Ramcharan et al. (2017). In addition to the 6 image classes implemented in Ramcharan et al. (2017), an additional nutrient deficiency class of 336 images was included in this work and examples of all image classes are shown in Figure 1.

For this study, three cassava disease experts reviewed images and agreed on classifications. Images were then annotated at Penn State University. Initially three different annotation styles were tested to identify class objects: (1) whole leaflet - object bounding boxes are drawn around leaflets with visible symptoms and boxes contain the leaf edges, (2) within leaflet - object bounding boxes are drawn around visible symptoms, inside of leaflets only, and do not contain leaf edges, and (3) combined inside and whole leaflet - annotation style (1) and (2) are combined with the same class labels for whole leaflet and within leaflet bounding boxes. Based on training results to 500 epochs on two 16Gb NVIDIA V100 GPUs reported in Table S1, the whole leaflet annotation style recorded the lowest overall loss and was selected to test on a mobile device in the field.

We selected three classes for detection in the field - cassava mosaic disease (CMD), cassava brown streak disease (CBSD), and green mite damage (CGM)[15]. For simplicity, CMD, CBSD and CGM are referred to collectively as disease in subsequent text. These diseases were selected as they are the major constraints to cassava production in sub-Saharan Africa [16,17]. Within each disease class, an IITA cassava disease expert identified 40 validation leaves which were split into 20 mild symptom leaves and 20 pronounced symptom leaves. Examples of the leaves in this dataset are shown in Figure 2. The average number of leaflets per leaf was 6, resulting in an average of 120 objects per disease/severity group. Where possible, all leaves for each disease were flagged on the same variety of cassava (Table S2). First, images of each of the 120 leaves were taken with a mobile device and model inference was run on a desktop to calculate performance metrics. Second, a mobile app video evaluation was conducted for the 120 leaves in overcast or cloudy conditions during the day. If weather conditions were sunny, an umbrella was used to shade leaves to obtain consistent light conditions across diseases. Leaves were also wiped to ensure surface was free of water and dirt. The CNN model was deployed on a Samsung Galaxy S5 Android device using the Tensorflow Demo App. The phone was held parallel to the leaf, at a distance such that all leaflets were visible in the frame. A screen capture was recorded for the first 10 seconds where the model was shown the leaf and bounding boxes were proposed. These videos were then downloaded and used to calculate the real world video performance metrics for the model.

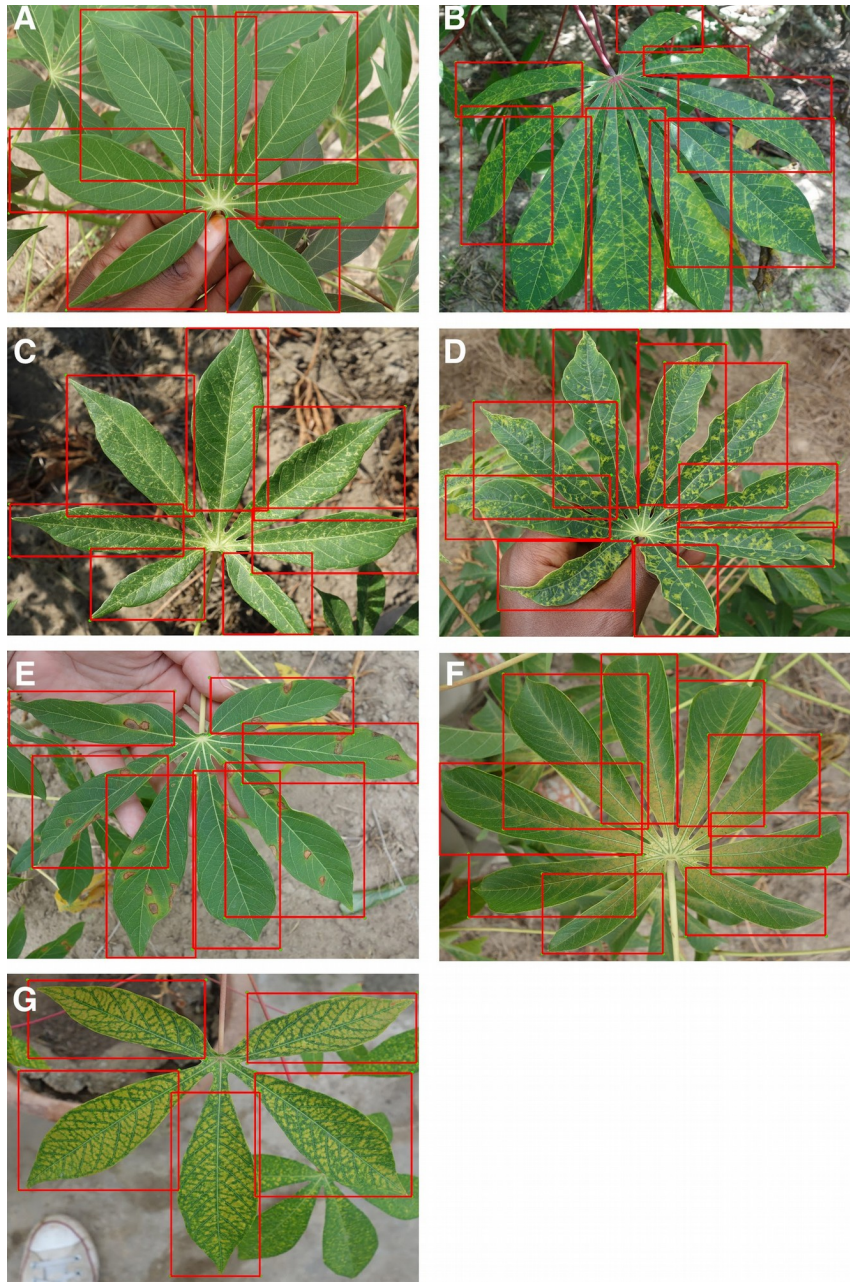

Figure 1 Examples of training images from 7 classes with leaflet annotations. Classes are (A) Healthy, (B) Brown streak disease, (C) Mosaic disease, (D) Green mite damage, (E) Red mite damage, (F) Brown leaf spot and (G) Nutrient Deficiency.

Table 1 Mean average precision of CNN model for real world images and video.

|  |  |  | CBSD | CMD | CGM |
|---|---|---|---|---|---|
|  | Severity | mAP | Class average precision | | |
| Test dataset | Pronounced | **94 ± 5.7** | 87.5 | 98.3 | 96.2 |
|  |  |  |  |  |  |
| Real world image | Pronounced | **91.9 ± 10** | 81.8 | 97.9 | 89.1 |
| Real world video | Pronounced | **89.6 ± 10** | 81.0 | 100 | 94.7 |
|  |  |  |  |  |  |
| Real world image | Mild | **75.0 ± 19** | 61.1 | 100 | 82.6 |
| Real world video | Mild | **81.2 ± 27** | 45.7 | 100 | 79.3 |
|  |  |  |  |  |  |
|  |  | mAR | Class average recall | | |
| Test dataset |  | **67.6 ± 4.7** | 62.7 | 68.2 | 72.0 |
|  |  |  |  |  |  |
| Real world image | Pronounced | **39.3 ± 10.9** | 32.5 | 33.6 | 51.9 |
| Real world video | Pronounced | **39.8 ± 19.9** | 21.0 | 25.0 | 57.3 |
|  |  |  |  |  |  |
| Real world image | Mild | **16.8 ± 10.9** | 21.1 | 4.40 | 25.0 |
| Real world video | Mild | **15.8 ± 10.3** | 11.3 | 8.50 | 27.6 |

Table 2 F-1 Scores for real world evaluation

| Test dataset | Severity | F-1 score |
|---|---|---|
| Test dataset | Pronounced | 0.79 |
|  |  |  |
| Real world image | Pronounced | 0.54 |
| Real world video | Pronounced | 0.48 |
|  |  |  |
| Real world image | Mild | 0.26 |
| Real world video | Mild | 0.25 |

Validation results for the model are provided in three ways and results are presented for the three disease classes studied in the field. First, precision and recall results with an 80-20 training-testing data split is reported in Table 1. The mean average precision (mAP) is the average across N classes of the true positive class labels divided by the total number of objects labeled as belonging to the positive class. The mean average recall (mAR) is defined as the average across N classes of the number of true positive class labels divided by the total number of ground truth positive class labels. These metrics were calculated assuming the cost of a false positive was equal to no predictions for a leaflet. For the disease classes of interest, the CNN detection model achieves 94 ± 5.7% (mean ± s.d.) mAP (67.6 ± 4.7% mAR) for the test dataset. Second, the results of the mAP and mAR evaluation using 120 images (comprising 742 leaflet "objects") of the field experimental leaves run on a desktop are reported. For pronounced symptomatic leaves, the model achieves 91.9 ± 10% mAP and 39.3 ± 10.9% mAR, while for mild symptomatic leaves, the model achieves 75.0 ± 19% mAP and 16.8 ± 10.9% mAR. Third, the results of the precision and recall evaluations from 120 screen capture videos (of the 742 leaflet objects) are reported. For pronounced symptomatic leaves, the model achieves 89.6 ± 10% mAP and 39.8 ± 19.9% mAR, while for mild symptomatic leaves, the model achieves 81.2 ± 27% mAP and 15.8 ± 10.3% mAR. These results show that the model maintains its average precision for pronounced symptoms in real world images and video and there is a small drop in performance for mild symptoms. With respect to precision the CNN model does slightly better on CMD symptoms and slightly worse on CBSD and CGM. The model recall is reduced by almost half its test dataset value in real world images and video of pronounced symptoms. This reduction is almost four times as large in mild symptom real world images and video.

The F-1 scores are also reported in Table 2. The F-1 score takes into account false positive and false negatives as it is a weighted average of precision and recall. The F-1 scores are 0.54 and 0.48 for pronounced symptoms in real world images and video respectively. For mild symptoms the F-1 scores are 0.26 and 0.25 for real world images and video respectively. The results show there is a 32% drop in F-1 score moving from the test dataset to pronounced symptoms in real world images (the closest data to the training data). For pronounced symptoms in real world video the F-1 score is reduced by 39%. Comparing the test dataset to real world mild symptoms the F-1 score drops by 67% for images and video. These results show there is a noticeable drop in CNN model performance from the test dataset to real world conditions with real world images performed slightly better than real world video for both pronounced and mild symptoms. Based on F-1 metrics, the CNN model does the best with CGM, followed by CMD, then CBSD.

$$\text{mAP} = \frac{1}{N}\sum_{n=1}^{N} \frac{\Sigma\, True\, Positive\, Detections}{\Sigma\, All\, Positive\, Detections} \quad (1)$$

$$\text{mAR} = \frac{1}{N}\sum_{n=1}^{N} \frac{\Sigma\, True\, Positive\, Detections}{\Sigma\, Ground\, Truth\, Positive\, Labels} \quad (2)$$

$$\text{F-1 score} = \frac{2 * Recall * Precision}{(Recall + Precision)} \quad (3)$$

We also calculated accuracies for the CNN model on the real world images and screen capture videos. The accuracy is calculated as the percent of the examples the model correctly detects i.e. the proportion of the observations where the predicted and ground truth annotations match.

Images and videos were reviewed by a cassava disease expert in order to calculate these metrics. Accuracy results for the CNN model run on 120 images (comprising 742 leaflet "objects") of the field experimental leaves are reported in Table 3. For pronounced symptomatic leaves, the model achieves 80.6 ± 4.10% accuracy, while for mild symptomatic leaves, the accuracy reduces to 43.2 ± 20.4%. The accuracy evaluation is then repeated using 120 screen capture videos of the mobile CNN model running real time in the field. For pronounced symptomatic leaves, the model achieves 70.4 ± 22.5% accuracy, while for mild symptomatic leaves, accuracy reduces to 29.4 ± 12.2%.

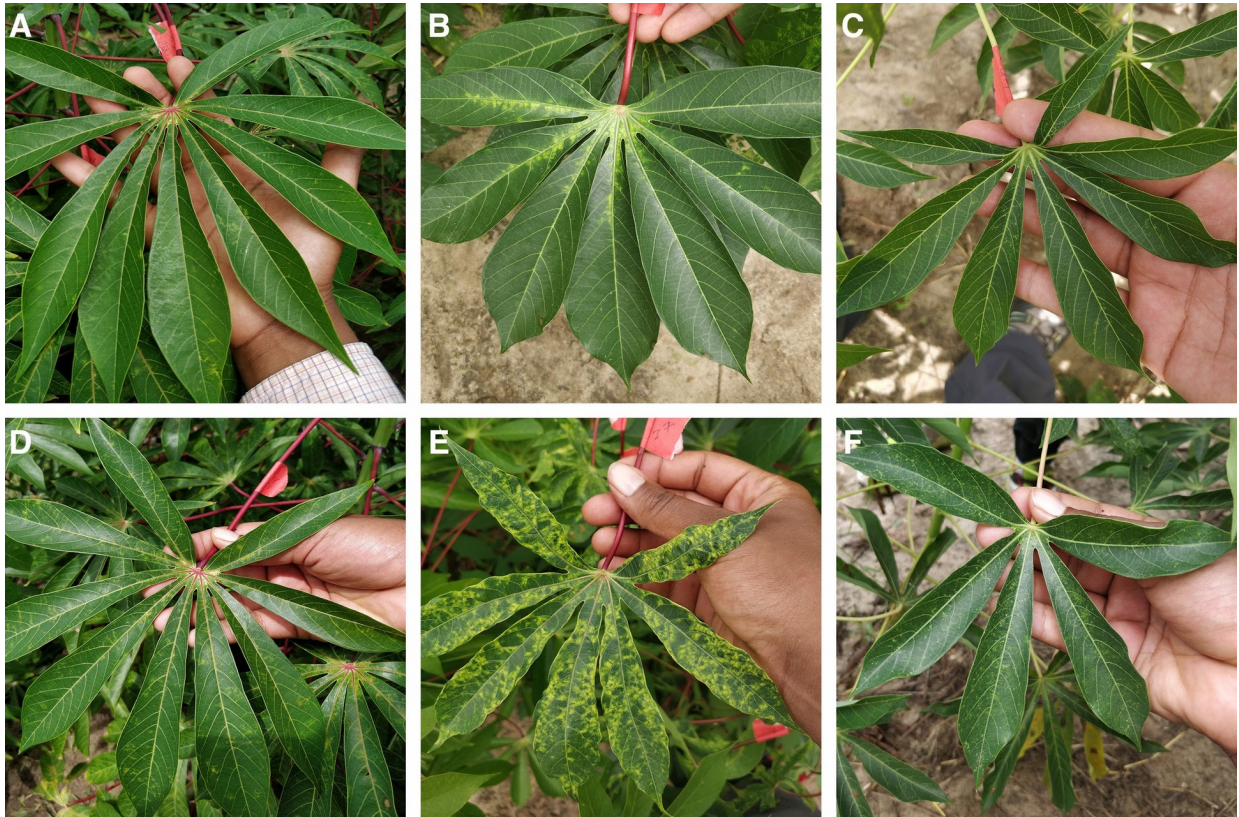

Figure 2 Examples images from field experiment showing mild symptoms (A), (B), (C) and pronounced symptoms (D), (E), (F) of CBSD, CMD, and CGM respectively.

Confusion matrices for the real world image and video model experiments give a more detailed analysis on how the model performance changes for different class/severity categories. In the confusion matrix plots in Figure 3, the rows correspond to the true class, and the column shows the model predicted class. The diagonal cells show the proportion (range 0-1) of the examples the model correctly detects i.e. the proportion of the observations where the predicted and ground truth annotations match. The off-diagonal cells show where the model made incorrect predictions.

Table 3.Accuracy results for the CNN model for real world images and video.

|  | Severity | Accuracy | CBSD | CMD | CGM |
|---|---|---|---|---|---|
|  |  |  | Class average accuracy | | |
| Real world image | Pronounced | **80.6 ± 4.10** | 76.1 | 83.9 | 81.7 |
| Real world video | Pronounced | **70.4 ± 22.5** | 45.9 | 90.3 | 74.0 |
|  |  |  |  |  |  |
| Real world image | Mild | **43.2 ± 20.4** | 61.1 | 21.1 | 47.5 |
| Real world video | Mild | **29.4 ± 12.2** | 23.9 | 20.8 | 43.4 |

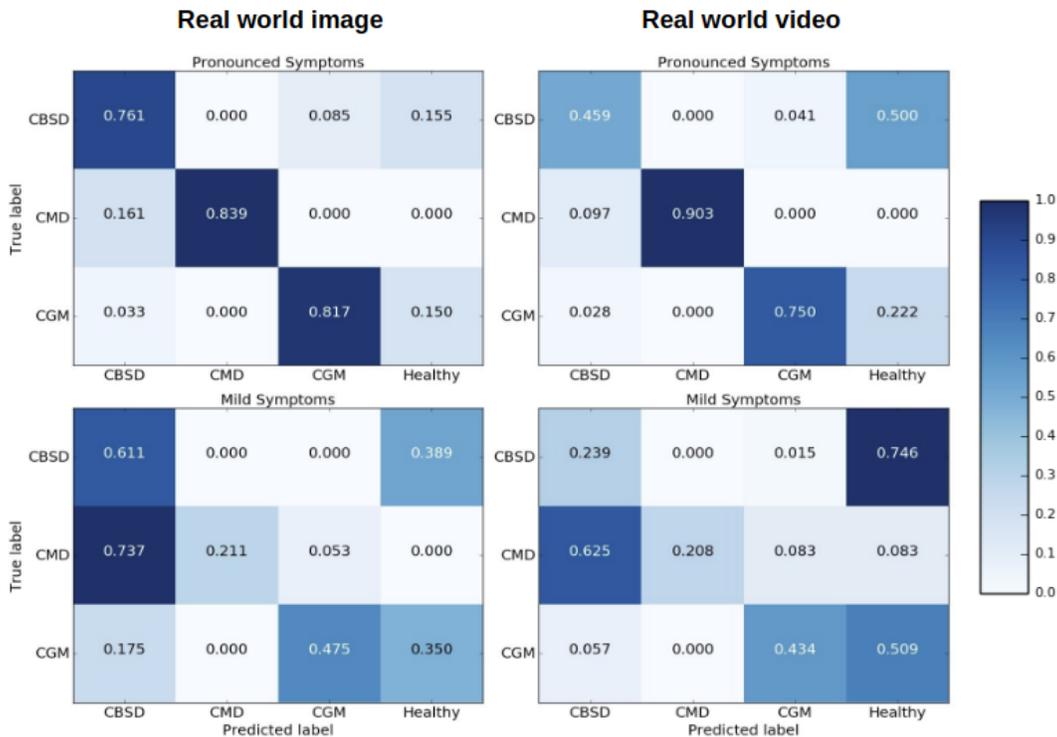

Figure 3 Confusion matrices for real world mobile images and real world mobile video.

The diagonal cells of the confusion matrices for leaves with pronounced symptoms show the model performs, as expected, much better where the symptoms are pronounced compared to leaves where the symptoms are mild. The biggest drop in accuracy going from pronounced symptoms to mild symptoms was for CMD for both image and video inferences of the model. This result can be due to the distortions of the leaf shape that occur for pronounced symptoms of CMD which are less obvious where symptoms are mild; the leaf distortion effect of disease does not occur for CBSD and only occurs in severe damage by CGM (not captured in this experiment). These results suggest that if symptoms change significantly during the different stages of infection, a model trained on one stage will be less reliable in detecting a different stage of infection of a disease. Comparing the performance of the model on images and video, the model accuracy is not significantly different for CMD and CGM classes but there is a surprising drop in accuracy for CBSD. There is a significant difference in CBSD accuracy for images (M= 0.73, SD = 0.10) and video (M=0.35, SD = 0.11); t(36) = 6.64, p = 0.0. This may be due to the subtlety of infection of CBSD. CBSD symptoms are not localized on an area of the leaf (unlike CMD and CGM). The contrast in color and patterns of CBSD symptoms is less pronounced than CMD and CGM. This could make CBSD symptoms more sensitive to motion blur and compression artifacts causing frame-to-frame variability, even though videos appear smooth to the eye[18]. The accuracy of the model could be improved with domain adaptation algorithms[19] to reduce this performance gap. The confusion matrices also show that the CNN model confuses mild symptoms with healthy leaves for symptoms that do not result in distortions in leaf shape. For CMD, the model was very effective in detecting severe symptoms, but very poor with mild symptoms. The leaf distortion that occurs in severe CMD infection makes identification of that class straightforward, but where symptoms are mild and leaf distortion was absent, there was a high level of confusion with CBSD (false positive rate is high for CBSD). This is unsurprising, as the difficulty of distinguishing between mild symptoms of these two diseases is a common problem faced in real-world field situations by cassava researchers.

In this study we evaluate the performance of a CNN model deployed offline in real time on a mobile device to detect foliar symptoms of cassava pests and diseases. Using the single-shot detector model, a CNN architecture optimized for mobile devices, we assess the performance of the model to detect pronounced and mild symptoms of 3 diseases classes. In both severities we see a decrease in F-1 score comparing the test dataset results to real world images and video. The decrease in performance was mostly due to the decrease in recall as the models overall maintained precision in real world conditions. Accuracy results also reflected the decrease in performance moving from real world images to video. The performance of the model in the mobile video changed depending on the difficulty of the visual diagnostic task. In order to obtain higher accuracy detections, there are a number of potential solutions - feed the CNN model images collected on a mobile device instead of real time video assessment, train on video images saved directly from the mobile app, or employ domain adaptation algorithms to improve performance. The CNN model also decreased in performance for mild symptoms, with accuracies decreasing more for diseases that change the leaf characteristics considerably during different stages of infection. In order to create a model that can detect mild symptoms of disease, images of mild infection are needed for model training. Collecting these images based solely on visual characteristics may be difficult as some groups of image classes e.g. viral diseases, may look similar to each other where infection is not severe. PCR tests may be needed to complement images to confirm symptoms. This study demonstrates the need to evaluate CNN performance in realistic operating conditions, with multiple performance metrics in order to validate models with data encountered in typical settings. Based on our results, we also recommend mobile CNN models to be used on the specific type of data with which it was trained until sufficient training examples exist from diverse data sources to better capture the diversity of data that occur in the real world.

# 2 Methods

## 2.1 Data Preprocessing

The cassava leaf dataset of JPEG images were taken with a Sony Cybershot 20.2-megapixel digital camera. Complete details of this dataset were previously reported in Ramcharan et al., 2017. For this study, IITA cassava experts extracted 2,415 images from the dataset based on the visibility of the most severe symptoms of each class. This was done under the assumption that the model would learn from the clearest examples of the symptoms and could then detect milder symptoms. The dataset totalled seven class labels as follows: three disease classes - cassava mosaic disease (CMD) (391 images), cassava brown streak disease (CBSD) (395 images), and brown leaf spot (BLS) (130 images), two mite damage classes - cassava green mite damage (GMD) (435 images), and red mite damage (RMD (351 images), and one nutritional deficiency class (NUTD) (336 images).

LabelImg[20], an open source graphical annotation tool for drawing and labeling object bounding boxes in images, was employed to draw ground truth bounding boxes and create corresponding xml files with stored xmin, xmax, ymin, ymax data for each ground truth box. Images and corresponding xml files were then converted to TFRecord files to be implemented in the Tensorflow environment. TFRecord files combine all images and annotations into one file, thereby reducing training time as it eliminates the need to open individual files. Each of the three models were then trained to 500 epochs.

## 2.2 CNN Model

We evaluated the performance of the CNN model built using standard precision metrics as well as a field-based independent evaluation on a mobile device. For the object detector model architecture, we selected the Single Shot Multibox (SSD) model with the MobileNet detector and classifier[21]. This model was used as it is one of the fastest object detection models available through Tensorflow[22]. An SSD model performs the tasks of object localization and object classification in a single forward pass - a key component in providing real time object recognition on a mobile device[11]. A pre-trained SSD model checkpoint trained on the COCO dataset (Common Objects in Context) was downloaded from Tensorflow's Detection Model Zoo (https://github.com/tensorflow/models/blob/master/research/object_detection/g3doc/detection_model_zoo.md) and transfer learning was employed to fine tune the model parameters. COCO is a large scale object detection, segmentation and captioning dataset comprised 330K images, 1.5 million object instances, and 80 object classes.

Each model was trained up to 500 epochs using a batch size of 15 on 2 NVIDIA Tesla V100 GPU's in Azure, the Microsoft cloud computing and storage platform. An 80-20 training-evaluation data split was used as this data partitioning scheme produced the best results for cassava disease classification[7]. The SSD model parameters were selected as follows: initial learning rate of 0.004, weighted sigmoid classification loss function, and weighted smooth L1 localization loss function between the predicted bounding box (*l*) and ground truth bounding box (*g*). These loss functions are computed based on default bounding boxes - a set of boxes with specified aspect ratios. The classification loss function measures the model's confidence in classifying pixels within a default bounding box into one class[11]. Localization loss measures the geometric distance between a default bounding box and the ground truth annotation bounding box. The overall loss function is a weighted combination of the classification loss (classif) and the localization loss (loc) (Eqn. 1) with the weight for the localization loss, $\alpha$, set to 1[21].

$$L(x, c, l, g) = \frac{1}{N} (L_{classif}(x, c) + \alpha L_{loc}(x, l, g))$$

(1)

The maximum number of hard mining examples was set to 3000, and the ratio between negative and positive examples was left at the default value of 3:1. The default box generator was applied to 6 different convolution layers with a minimum and maximum scale of 0.2 and 0.95 respectively. The default boxes were generated with fixed aspect ratios 1.0, 2.0, 0.5, 3.0, and 0.333. The complete details of the SSD model design principles are provided in Liu et al (2016)[19]. Finally, in order to perform real time inference on a mobile device, images were resized to 300 x 300 pixels before being fed into the network.

## 2.3 Annotation Experiment Results

The mean average precision (mAP) across all classes as well as for each class for the annotation experiment are reported in Table S1. These metrics were then repeated for the in-field model assessment of the CNN model for annotation style 1 (whole leaflet).

Table S1 Mean average precision for 3 annotation styles studied.

| Annotation Style | mAP | Class average precision | | |
|---|---|---|---|---|
| | | CBSD | CMD | CGM |
| 1 - whole leaflet | **71.9** | 66.4 | 66.0 | 83.2 |
| 2 - within leaflet | **13.6** | 17.1 | 11.9 | 11.8 |
| 3 - combined whole + within leaflet | **10.9** | 15.0 | 9.33 | 8.27 |

# Competing Interests

The authors declare no competing financial interests.

# Acknowledgements

We would like to thank the following undergraduate students for assistance with annotating images and reviewing images and video for the CNN model evaluation: Ryan Ondocin, Jonathan Meyer, Jordan Moreno, Holly Bartell, Gracie Pilato, and Gabrielle Perseghin. This work was supported by funding from the Huck Institutes at Penn State University and the CGIAR through their INSPIRE award.

# Author Contributions

A.M.R. conceptualized experiment, selected algorithm, collected and analyzed data, and wrote manuscript. P.M. trained algorithms, collected and analyzed data, and wrote manuscript. K.B. analyzed data. N.M. and M.N. collected data. L.M., J.L. and D.H. are project supervisors.